\pdfoutput=1
\documentclass[12pt]{iopart}
\usepackage[T1]{fontenc} 
\usepackage{graphicx}
\usepackage{booktabs} 
\usepackage{amssymb}

\usepackage{caption} 
\usepackage{hyperref} 

\begin{document}

\title[Causal impact of winter wheat on SOC]{Winter wheat improves soil carbon sequestration with reduced gains in warmer climates}

\author{Georgios Giannarakis$^{1,*}$, Vasileios Sitokonstantinou$^{2,*}$, Dimitrios Bormpoudakis$^1$, Ilias Tsoumas$^{1,2}$, Nikiforos Samarinas$^{4,5}$, Gustau Camps-Valls$^3$ and Charalampos Kontoes$^1$}

\address{$^1$ Beyond Center, National Observatory of Athens, Athens, Greece}
\address{$^2$ Artificial Intelligence Group, Wageningen University \& Research, The Netherlands}
\address{$^3$ Image Processing Laboratory (IPL), Universitat de València, València, Spain}
\address{$^4$ Laboratory of Hydraulic Works and Environmental Management, Dep. of Rural and Surveying Engineering, Aristotle University of Thessaloniki, Thessaloniki, Greece} 
\address{$^5$ Interbalkan Environment Center, 18 Loutron Str., Lagadas, Greece}

\vspace{10pt}
\begin{indented}
\item[] $^*$These authors contributed equally to this work.
\item[] October 2025
\end{indented}

\ead{vassilis.sitokonstantinou@wur.nl} 



\begin{abstract}
Sustainable agricultural intensification requires a precise understanding of how land use decisions influence soil organic carbon (SOC) sequestration under varying environmental conditions. We apply causal machine learning to evaluate the heterogeneous impact of winter wheat-based rotations on SOC at the field-level across Lithuania. Using a nationwide dataset collected over five years (2018–2022), including crop types, management practices, soil properties, and environmental conditions, we estimate how the effects of winter wheat-based rotation vary across space and climate. Our results indicate that winter wheat generally enhances SOC levels, but its effectiveness is modulated by climate. Cooler regions experience higher SOC gains, whereas warmer regions show diminished or even negative effects. Future climate projections suggest that the potential for winter wheat carbon sequestration may decline under high-emission scenarios. These findings highlight the importance of localized agricultural strategies that account for climate variability to maximize SOC sequestration and ensure long-term soil sustainability.

\end{abstract}

%
%
%
%
%
\section{Introduction}

Agriculture plays a vital role in the global carbon cycle, with its total greenhouse gas emissions comprising 10-12\% \cite{ipcc2022wg2} of total emissions with a hidden cost of up to 1.26 trillion dollars globally \cite{fao2024sofa}. At the same time, croplands hold 83 PgC with an additional storage capacity between 29 to 65 PgC \cite{padarian2022soil}, with the potential to offset 4\% of average annual global greenhouse gas emissions over the rest of the century \cite{c83ea859a5504a7d9b5eda66a8ea0838}. Although it is highly vulnerable to climate variability, improved practices such as optimized crop rotations and conservation tillage have been shown to boost soil organic carbon (SOC) levels respectively by 3.3 \% \cite{li2024crop} and 4.5–23.7 \% \cite{meng2024meta} in some regions. SOC, a key indicator of soil health, directly influences productivity, water retention, and long-term sustainability \cite{Toensmeier2016}. In fact, increasing SOC can substantially enhance agricultural resilience and contribute to mitigating climate change. However, optimizing SOC levels requires a precise understanding of how different cropping systems affect carbon storage under varying environmental conditions \cite{giannarakis2022personalizing}.

One of the key challenges in SOC management is that the impact of farming decisions on SOC is not uniform; it depends on soil properties, crop selection, climate conditions, and management intensity. Policies aiming to increase SOC often rely on generalized recommendations, but these one-size-fits-all approaches fail to account for spatial variability \cite{li2023,behrendt2024}. As a result, site-specific insights are needed to tailor agricultural strategies that balance productivity, climate mitigation, and long-term sustainability.

Winter wheat-based crop rotations provide an important case study for assessing SOC dynamics. Winter wheat is a widely cultivated cereal with known agronomic and economic benefits, yet its long-term effects on SOC remain context-dependent. While intensive wheat cultivation may deplete soil nutrients, multi-year wheat-based rotations have been associated with increased carbon inputs through root biomass and crop residues, potentially enhancing SOC storage \cite{persson2008long, alhameid2017soil,studdert2000crop}. Assessing these effects at fine spatial scales is essential for understanding whether and where such rotations contribute to sustainable intensification \cite{halvorson2012crop}.

Recent progress in big Earth data and Artificial Intelligence (AI) has improved how we monitor agriculture. Remote sensing combined with machine learning is widely used for tasks like crop classification, yield estimation, and disease detection \cite{liakos2018machine}. However, most of these models are designed to find patterns and make accurate predictions, not to understand the underlying causes of observed outcomes \cite{giannarakis2022towards}. For example, a model might predict SOC levels from satellite and soil data, but it cannot tell whether a specific farming practice caused SOC to increase or whether the change is due to other factors like climate or soil properties \cite{sitokonstantinou2024causal}. Answering such questions requires causal inference methods that move beyond correlations and explicitly model the processes driving observed changes \cite{cohrs2024causal,CampsValls23physcausaldiscovery,runge2023causal}.

Causal machine learning applied to observational data offers a promising way to estimate the effects of farming practices across space. Recent studies have explored its use for evaluating impacts on agroecosystem services \cite{deines_satellites_2019, giannarakis2022towards, giannarakis2023understanding}, highlighting its potential. However, these approaches often lack the field-level spatial resolution needed for decision-making in precision agriculture \cite{stetter2022using}, or they do not account for long-term effects under changing climate conditions \cite{smith2004long}.

In this study, we employ a data-driven approach to investigate the impact of winter wheat-based rotations on SOC across fields in Lithuania. We use a method called Double Machine Learning (DML) \cite{chernozhukov2018double,Cohrs23dml}, which helps us sort out the true impact of these rotations by separating it from other factors. This approach enables us to calculate the conditional average treatment effect, which measures the average impact on SOC for each field under specific conditions by combining detailed data on crops, soil, and the environment. With our analysis, we enable:
\begin{itemize}
\item \textbf{Personalized, data-driven recommendations for sustainable intensification}: By quantifying localized (field-level) treatment effects, we provide evidence-based guidance for farmers and policymakers on where winter wheat-based rotations enhance SOC and where alternative strategies may be needed.

\item \textbf{Assessment of climate sensitivity:} We analyze how key environmental factors, such as temperature and precipitation, modulate the impact of winter wheat-based rotation on SOC, identifying environmental thresholds where sequestration benefits diminish.

\item \textbf{Forward-looking climate scenarios:} By incorporating climate projections, we assess how SOC sequestration potential may evolve under future conditions, informing long-term sustainable land management.

\end{itemize}

\section{Data and Methods}

Our study examines the impact of winter wheat-based rotation (treatment) on SOC (outcome) in Lithuania from 2018 to 2022. Table \ref{tab:variables} summarizes the variables used for this analysis. A detailed description of the data and sources is provided in Sec. \ref{data}.

\subsection{Study area}

Our study captures the entire territory of Lithuania, located in northern Europe along the southeastern coast of the Baltic Sea. Lithuania has a humid continental climate characterized by moderate to heavy precipitation and significant seasonal temperature variations. It belongs to the cool, temperate and moist IPCC climate region \cite{eggleston20062006}. Summers are generally warm, with average temperatures around 17°C, while winters can be cold, with temperatures often dipping below freezing. This climate supports various agricultural activities, with about 50\% of the land being arable, which plays a key role in the national economy \cite{armolaitis2022evaluation}.

Agriculture in Lithuania is diverse, with cereals being a significant component. Winter wheat (Triticum aestivum), in particular, is a widely cultivated staple crop throughout the country. The main regions for winter wheat-based rotation include central and northern Lithuania, where soil conditions and climate favor its growth. This crop is typically sown in the autumn, is hung over a blanket of snow, and is harvested in the late summer. Winter wheat has the potential to improve SOC levels, which is essential for soil health and fertility. Winter wheat cultivation contributes to SOC by adding root biomass and residues that remain after harvest. Over time, these organic materials decompose, improving soil structure, water retention, and nutrient availability. Studies have shown that continuous winter wheat cultivation can gradually increase SOC, which is critical for combating soil degradation \cite{smith2008impact}. 

\subsection{Data}\label{data}

In this subsection, we describe the datasets used to estimate the effect of winter wheat-based rotations on SOC content, including data derived from crop type maps (i.e., cultivated crop types and crop management), soil data (SOC and clay content), climate and environmental data (see Table \ref{tab:variables}). 

\paragraph{Data derived from crop type maps}

We define winter wheat-based rotations (treatment) as fields where winter wheat was grown in at least three out of the five years of the study period. This information comes from the Lithuanian Land Parcel Identification System (LPIS), a national geospatial database that includes field boundaries, declared crop types, and whether farming is organic or conventional. Using this data, we also created a binary organic farming indicator (\textit{eco}), which equals 1 if a field was declared organic in at least one of the five years, and 0 otherwise. We define \textit{crop rotation} as the number of different crop types cultivated in a field during the study period. This variable is important because it can influence both the treatment (fields with diverse rotations are less likely to have repeated winter wheat) and the outcome (SOC). The effect of crop rotation on SOC is well supported in the literature \cite{west2002soil}. Finally, we include the geographic coordinates (\textit{geo}) of each field's centroid, since location can strongly affect both crop growth and soil conditions.

\paragraph{Soil data}\label{soil}

This study used national-scale maps of \textit{SOC} and \textit{clay} content at field-level resolution, produced by Samarinas et al. \cite{samarinas2023soil} for 2020–2022. These maps were generated using Sentinel-2 imagery and machine learning models. An important characteristic of these data is that the annual SOC layers are not fully independent. Before SOC and clay predictions are made, the study produces bare-soil reflectance composites, which rely on a three-year median of annual bare-soil detections. This multi-year window is necessary in northern European croplands, where cloud cover, snow and cropping cycles limit the availability of bare-soil pixels in any single year. As a result, consecutive SOC layers share underlying information and are highly autocorrelated. This design yields more complete and stable SOC estimates but limits the usefulness of short-term year-to-year changes. After compositing, a convolutional neural network trained on the LUCAS topsoil dataset predicts SOC and clay content, with reported uncertainties. For our analysis, we averaged SOC and clay values over 2020–2022 to obtain stable field-level estimates. SOC serves as the outcome variable and clay is included as a covariate due to its influence on soil physical structure and carbon stabilization processes \cite{matus2021fine}.

\paragraph{Environmental data}\label{env}

To analyze the effects of winter wheat-based rotations on SOC, we identified a set of relevant variables based on prior research and a thorough understanding of the study context (see Sec. \ref{method}). Land-based weather and climate variables data were retrieved from the ERA5 database for each field, including parameters such as ambient temperature (\textit{t2m}), total precipitation (\textit{tp}), wind speed (\textit{u10}), snow cover (\textit{snowc}), soil temperature (\textit{stl}), soil water content (\textit{swvl}), and solar radiation (\textit{ssro}) \cite{hersbach2020era5}. These variables capture key climatic conditions that influence both SOC and crop growth. By incorporating them into our analysis, we aim to understand how winter wheat-based rotations affect SOC under varying climatic conditions.


\paragraph{Climate projections} \label{clim}

To understand the evolving impact of climate change on winter wheat-based rotations, we utilized climate projection data from CMIP6 for Shared Socioeconomic Pathways (SSPs) 1-2.6, 2-4.5, and 5-8.5 scenarios. SSPs are socioeconomic trajectories describing potential global development pathways, including population growth, economic development, technological progress and policy decisions \cite{riahi2017shared}. SSP1-2.6 represents a sustainability-focused world, where global cooperation leads to low emissions and strong climate mitigation efforts, limiting global warming to well below 2°C. SSP2-4.5 reflects a middle-of-the-road scenario, where development trends follow historical patterns, resulting in moderate emissions. SSP5-8.5 assumes rapid fossil-fuel-driven economic growth, with high energy demand and minimal climate policies, leading to a high-emission, business-as-usual trajectory. These scenarios are widely used in conjunction with climate models to simulate future climate conditions, including changes in temperature, precipitation and extreme weather patterns.

This study used monthly temperature and precipitation projections from 2024 to 2100 for these three SSP scenarios, supplemented with historical climate data for comparison. By examining 2030–2060 and 2070–2100 projections, we assess how climate change may alter the relationship between winter wheat-based rotations and SOC content. We achieve this by performing inference using the trained causal model on the climate projection data. 

\paragraph{Handling geospatial data} 

Geospatial datasets were preprocessed to minimize inconsistencies, including filtering out fields with high prediction uncertainty for SOC and clay content, correcting implausible values and aligning all datasets to a common coordinate system and grid resolution. Some covariates, such as climate variables from ERA5, are available only at a coarser resolution ($\sim$9 km) than field-level SOC data. In these cases, we assigned the grid-level values to all fields within the pixel. This assumption is justified because SOC dynamics respond primarily to broad-scale and cumulative climatic conditions (e.g., mean seasonal temperature and precipitation) rather than highly localized microclimate variations. In Lithuania, this approach is further supported by the country’s low relief and smooth climatic gradients: seasonal temperature and precipitation fields vary coherently at scales $\geq$10 km, with sharp microclimatic discontinuities being rare outside the immediate coastal belt and some of the uplands. We expect sub-grid variability within $\sim$9 km to be driven more by hydrology than by baseline seasonal climate, which are treated explicitly elsewhere in the model (see inclusion of deep soil temperature, soil moisture content, and subsurface runoff). Thus, while this approach may introduce some aggregation error, the ERA5 resolution adequately captures the main climatic gradients relevant for our analysis. Despite controlling for observed confounders, nearby fields may still exhibit similar SOC changes due to spatial autocorrelation, meaning that clusters of apparently significant effects could partly reflect spatial clustering rather than independent treatment effects. Spatial resolution could also affect model performance: finer resolution captures local heterogeneity but may increase noise, while coarser resolution smooths variability and may underestimate localized effects.

\subsection{Methodology}\label{method}

We frame the problem of estimating the impact of winter wheat-based rotation on SOC at the field level as a Conditional Average Treatment Effect (CATE) estimation task using the Potential Outcomes framework \cite{rubin2005causal}. Let \( T \in \{0, 1\} \) denote a binary treatment indicating whether a field has been under winter wheat cultivation for more than three years. The corresponding potential outcome is \( Y(T) \), representing the SOC level under treatment \( T \). Our goal is to estimate \( \theta(x) = \mathbb{E}[Y(1) - Y(0) \mid X = x], \) which captures how the effect of winter wheat on SOC varies depending on field characteristics \( X \).

We distinguish between two sets of covariates: the heterogeneity vector \( X \) and the control vector \( W \). The vector \( X \) includes variables that are expected to moderate the effect of winter wheat on SOC and thus explain heterogeneity in treatment effects. Specifically, \( X \) includes 3-year average clay content (\textit{clay}), farming method (\textit{eco}), mean temperature (\textit{t2m}), and precipitation (\textit{tp}). These variables represent key soil and climate conditions known to affect SOC dynamics \cite{paustian2016climate, west2002soil}. In contrast, the control vector \( W \) includes variables that may confound the relationship between treatment and outcome but are not assumed to influence how the treatment effect varies. These include geolocation (\textit{geo}), crop rotation diversity (\textit{rotation}), and several ERA5-derived variables: wind speed (\textit{u10}) \cite{yan2005losses,easson1993effects}, snow cover (\textit{snowc}) \cite{juurakko2021cold}, deep soil temperature (\textit{stl}) \cite{davidson2000soil}, soil moisture content (\textit{swvl}) \cite{azlan2012correlation}, and subsurface runoff (\textit{ssro}) \cite{bechmann2021soil}. These variables are associated with both winter wheat adoption and SOC levels but are not expected to modulate the treatment effect. ERA5 variables are treated as stable at the field level, given their spatial resolution (9×9 km).

To estimate \( \theta(x) \), we employ DML \cite{chernozhukov2018double}, which allows for confounding control and unbiased estimation in high-dimensional settings with complex, potentially non-linear relationships among covariates. This makes DML particularly suitable for geospatial and environmental datasets, where variables such as soil properties, climate and management interact in complex ways \cite{gilbert2021causal}. In the first stage of DML, machine learning models (here Random Forests) estimate two nuisance functions: the outcome model \( \hat{Y}(W, X) \), predicting SOC, and the treatment model \( \hat{T}(W, X) \), predicting treatment assignment. In the second stage, residuals from these models, i.e., the portions of \( Y \) and \( T \) not explained by the covariates, are regressed against each other. This residual-on-residual regression isolates the variation in \( Y \) that can be causally attributed to \( T \), after adjusting for confounding. Under the assumption that all relevant confounders are observed, this procedure yields an unbiased estimate of the CATE. 

\begin{table}[htbp]
\centering
\begin{tabular}{c|l|l|l}
\toprule
\textbf{Variables} & \multicolumn{1}{c}{\textbf{Description}}          & \textbf{Units} & \textbf{Type} \\
\midrule
wheat              & winter wheat-based rotations for $>3$ years           & binary         & Treatment     \\
\midrule
SOC                & soil organic carbon content                       & g C/kg         & Outcome       \\
\midrule
clay               & three year average clay content                   & \%             & X             \\
rotation           & number of different crop types within the 5 years & 1-5            & W             \\
geo                & latitude and longitude of field centroid           & degrees        & W             \\
eco                & organic or conventional farming                   & binary         & X             \\
t2m                & Temperature at 2 meters                               & K              & X             \\
tp                 & Total precipitation                               & m              & X             \\
u10                & 10 metre U wind component                         & m s$^{-1}$     & W             \\
snowc              & Snow cover                                        & \%             & W             \\
stl               & Soil temperature level 4                          & K              & W             \\
swvl              & Volumetric soil water layer 4                     & m$^3$ m$^{-3}$ & W             \\
ssro               & Subsurface runoff                                 & m              & W            \\
\bottomrule
\end{tabular}
\caption{Variables used in the analysis, including their descriptions, units, and assigned roles as either heterogeneity features (\( X \)) or control variables (\( W \)).}

\label{tab:variables}
\end{table}

\subsection{Unobserved confounders and omitted variables}

Other unobserved factors may influence both the adoption of winter wheat and SOC levels, potentially introducing omitted variable bias in propensity score and treatment effect estimates. Examples include soil properties (e.g., pH, texture, initial organic matter), nutrient availability, previous land use, crop yields, and management practices such as fertilization, pesticide use, irrigation, residue management, and tillage. Many of these are known to affect SOC positively or negatively. For instance, high initial organic matter, effective nutrient and residue management, and proper irrigation tend to enhance SOC, while soil compaction, steep slopes, and excessive agrochemical use can reduce it \cite{jing2024whether,de2019long}. High initial SOC in particular is a potential source of heterogeneity \cite{anand2025classifying,basso2025multi}. We note that initial (prior to 2018) SOC levels are not available to this study, as our field-level soil data coincide with the treatment period and will bias results if included in the model as a control. Studies with access to pre-treatment SOC data would be well placed to explore this question further.

To partially mitigate the influence of such unobserved heterogeneity, our model includes the geographical location of each field. Location implicitly captures spatially structured differences in management intensity, soil characteristics, and climatic conditions that are often correlated with agronomic practices such as fertilization, residue management, and tillage. While not a perfect substitute for direct observations, it is a useful proxy for regionally consistent management and environmental conditions.

Despite the possibility that some important factors were not observed or included in our model, we found that the treatment and control groups were well balanced in terms of the characteristics we could measure. Specifically, their predicted probabilities of receiving the treatment (propensity scores, shown in Supplementary Material) showed substantial overlap, supporting the internal validity of our analysis. This balance suggests that, conditional on observed covariates, the two groups are comparable. Nevertheless, propensity score methods can only account for observed confounders; unmeasured factors may still bias estimates. Large-scale randomized or quasi-natural experiments that directly account for such unobserved factors would be valuable to further test the robustness of our findings.

\section{Results and discussion}

We took several preprocessing steps to ensure the reliability of the data. First, we applied uncertainty filtering to remove fields with high prediction uncertainty for SOC and clay content. Since uncertainty values ranged from 0 (lowest) to 1 (highest), we excluded fields with uncertainty values above the third quartile, specifically fields with SOC uncertainty over $0.51$ and clay content uncertainty over $0.69$. After filtering, we calculated each field's average SOC and clay content over 2020–2022 to obtain stable, representative estimates.
Additionally, to maintain consistency in land cover, we excluded fields not continuously classified as arable land across all five years in the dataset. We acknowledge that the underlying AI-predicted maps have a moderate R$^2$ of 0.51. While this is lower than ideal, it is comparable to similar large-scale SOC mapping studies in the literature (see \cite{lima_comparing_2019}), and the extensive spatial coverage combined with uncertainty filtering allows meaningful inference of large-scale SOC trends.

Next, we applied propensity score filtering to ensure that the treatment and control groups were comparable and to reduce bias in our causal effect estimates. The propensity scores represent the probability of a field adopting a winter wheat-based rotation based on its observed characteristics. To estimate these scores, we standardized the combined feature set \(X + W\) using a standard scaler. We trained a Gradient Boosting classifier with 3-fold cross-validation to predict the treatment variable \(T\). To ensure sufficient overlap and avoid extrapolation in regions with little comparability, we excluded fields with propensity scores outside the range of $0.2$ to $0.8$ (see Figure A1 in Supplementary Material). This trimming step removed units for which treatment assignment was either highly predictable (scores close to 1) or highly unlikely (scores close to 0), focusing our analysis on the region of common support. After these preprocessing steps, 70,000 fields remained for analysis.

Then, we estimated the CATE using the Linear DML model, which was selected for its interpretability. The outcome variable \(Y\) and the treatment variable \(T\), in the first stage of DML, were modeled using Random Forest to capture complex, non-linear relationships within the data. Finally, we performed inference on a hold-out dataset to robustly and impartially evaluate our estimated causal effects.

\subsection{Causal effect of wheat on SOC}

Our results indicate that winter wheat-based rotations in Lithuania are linked to significantly higher SOC content, with an estimated Average Treatment Effect (ATE) of $+0.44$ g/kg (see Table \ref{table}). In other words, after adjusting for confounding factors, fields adopting this rotation practice exhibit increased SOC levels. This estimate is derived from the CATE model's intercept term, which is equivalent to the ATE because all covariates ($X$) were centered to their mean of zero during preprocessing.

Winter wheat-based rotations work by extending the period of active plant growth into the winter, contributing to longer intervals of continuous soil coverage and below‐ground carbon inputs. This finding is especially relevant in Lithuania, where approximately half of the country’s agricultural land remains bare throughout the winter \cite{eurostat2020}. Local studies support our results: Liaudanskiene et al. \cite{liaudanskiene2011} report that increasing the proportion of wintering crops, including winter wheat, in a rotation has a positive significant influence on the humification degree of soil organic matter in the plow layer. Similarly, Slepetiene et al. \cite{slepetiene2010} found that increasing the share of winter crops (75–100\%) in a rotation boosts humus content in the topsoil, showcasing how winter wheat and similar species such as oilseed rape help conserve soil organic matter. While agricultural management practices such as organic inputs \cite{majumder2008} or tillage \cite{saldukaite2022} can modulate these outcomes, the basic function of winter wheat, and other winter crops, is the active cover of the soil and deposition of root biomass for a greater portion of the year \cite{patra2022,zhang2024}.

Broadly, the Lithuanian findings support the idea that functional crop types matter more than sheer crop-species diversity. Wooliver et al. showed that particular crop‐species thresholds for maximizing soil microbial diversity and SOC do not exist \cite{wooliver2022}. Mpeketula et al. \cite{mpeketula2019} observed that adding more crops does not necessarily significantly improve SOC. In contrast, winter crops, especially in rotation, ensure extended ground cover as plant cover duration is a key driver of below‐ground carbon processes \cite{garland2021crop}. In summary, including winter crops in Lithuanian rotations, even in as many as four out of five years, is a practical and effective strategy to enhance humification, improve soil fertility, and potentially support strong yields.

\subsection{Effect heterogeneity}

\begin{figure}[!ht]
\centering
\includegraphics[width=0.9\columnwidth]{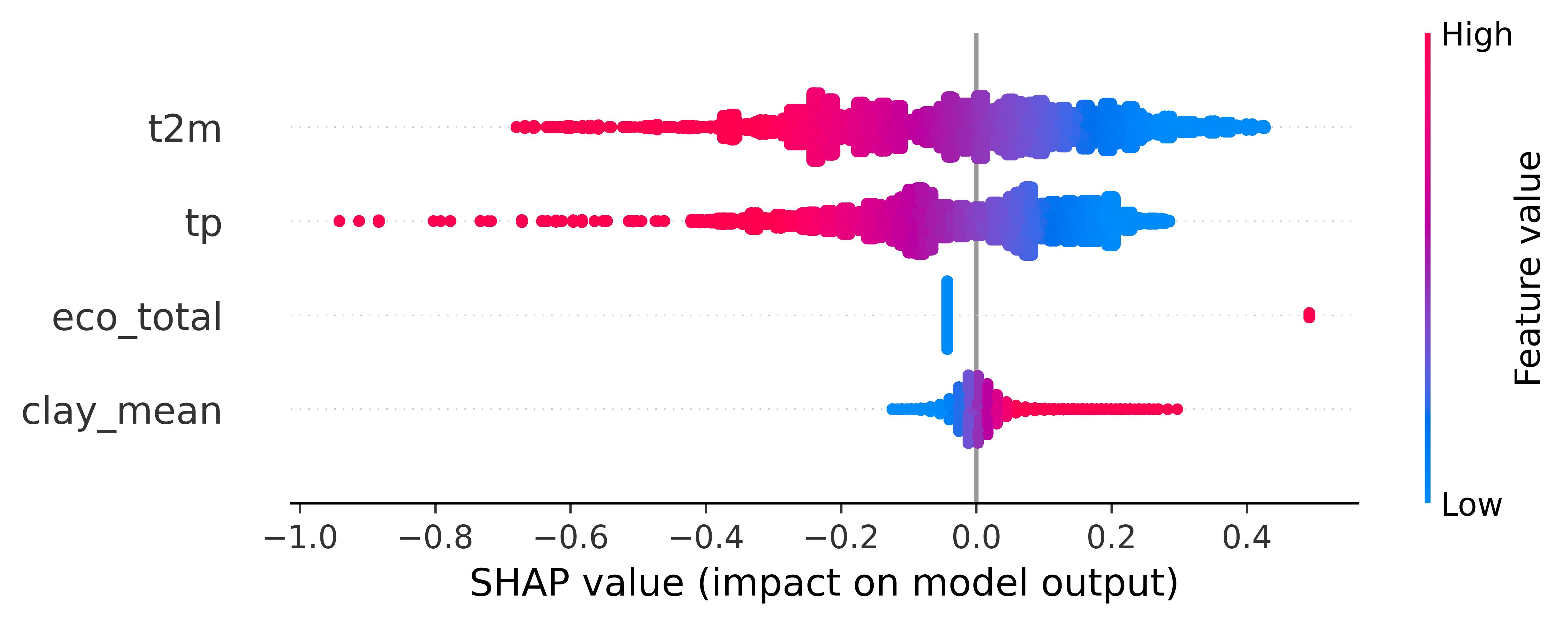}
\caption{SHAP summary plot depicting the relative importance of various features in predicting the estimated CATE of winter-wheat–based rotation on SOC content. The horizontal axis represents SHAP values, indicating the magnitude and direction of each feature’s contribution. Temperature (t2m) and total precipitation (tp) emerge as influential modulators of treatment effects. Organic farming (eco) also shows a relatively large estimated influence; however, this effect is characterized by high uncertainty due to the small number of available samples (as reflected in its standard error). Clay content shows comparatively limited contribution.}
\label{fig:importance}
\end{figure}

We found that the impact of winter wheat-based rotations at the field-level on SOC content is influenced by local environmental factors. Higher temperatures were associated with a significant decrease in SOC ($-0.19$ g/kg), while total precipitation also showed a negative association ($-0.16$ g/kg), although this effect was not statistically significant at the 95\% confidence level. Nonetheless, it provides a directional insight, consistent with earlier studies \cite{hontoria1999, jobbagy2000}. In contrast, clay content and organic farming (\textit{eco}) status did not show statistically significant effects (see Table \ref{table} and Fig. \ref{fig:importance}).

\begin{table}[htbp]
    \centering
    \begin{tabular}{lccccc}
        \toprule
        Coefficient & Point Estimate & Std. Error & Z-score & P-value & 95\% CI \\
        \midrule
        clay   & 0.033  & 0.060  & 0.545  & 0.586  & [-0.085, 0.151] \\
        eco  & 0.535  & 0.654  & 0.819  & 0.413  & [-0.746, 1.816] \\
        t2m        & -0.192 & 0.086  & -2.217 & 0.027  & [-0.361, -0.022] \\
        tp         & -0.159 & 0.094  & -1.683 & 0.092  & [-0.344, 0.026] \\
        \midrule
        CATE intercept & 0.439 & 0.089 & 4.914 & 0.000 & [0.264, 0.614] \\
        \bottomrule
    \end{tabular}
    \caption{Estimated coefficients for key covariates affecting the CATE of winter wheat-based rotations on SOC content. The table reports point estimates, standard errors, Z-scores, p-values, and 95\% confidence intervals for each predictor.}
    \label{table}
\end{table}

\begin{figure}[!ht]
\centering
\includegraphics[width=0.8\columnwidth]{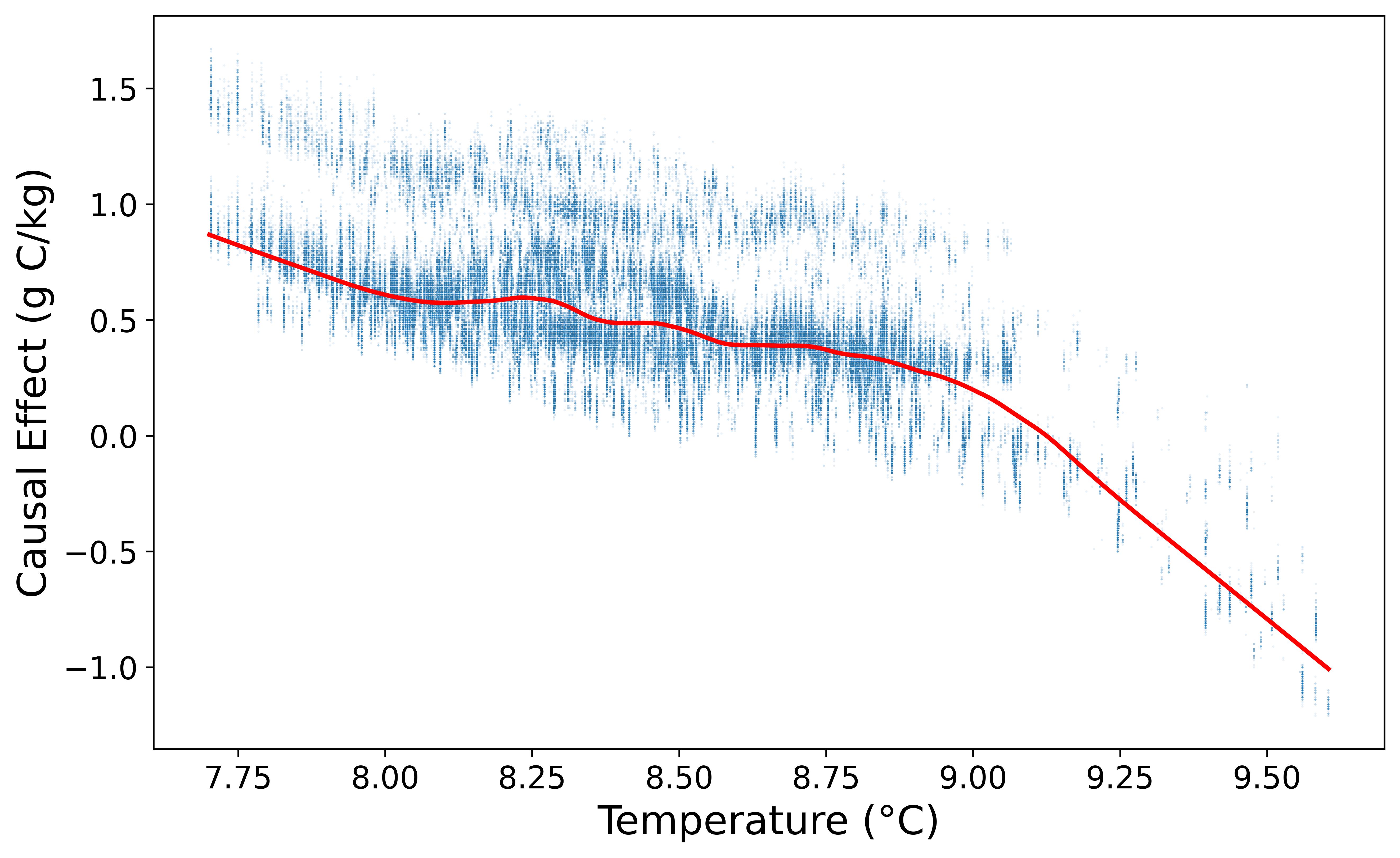}
\caption{Causal effects as a function of temperature: Nonlinear relationship (LOWESS smoothed) between mean annual temperature and SOC content. Results indicate a significant negative association: higher temperatures correspond to lower SOC retention.}
\label{fig:temp}
\end{figure}

We found that temperature is nonlinearly negatively related to SOC impacts, as illustrated in Fig.\ref{fig:temp}. This finding aligns with previous studies showing that SOC tends to decrease with increasing temperatures in a nonlinear manner \cite{kirschbaum1995, zhao2013}. Lower temperatures are associated with higher SOC gains ($0.47$), with even greater gains observed in organic fields. In contrast, high temperatures combined with high precipitation drive the treatment effect toward zero, suggesting minimal SOC benefits under these conditions. Complementary to the SHAP analysis discussed in Figure \ref{fig:importance}, Supplementary Material include a decision tree model which illustrates how temperature and precipitation influence the estimated treatment effect as well as Partial Dependence Plots for key covariates.

\begin{figure}[!ht]
\centering
\includegraphics[width=0.8\columnwidth]{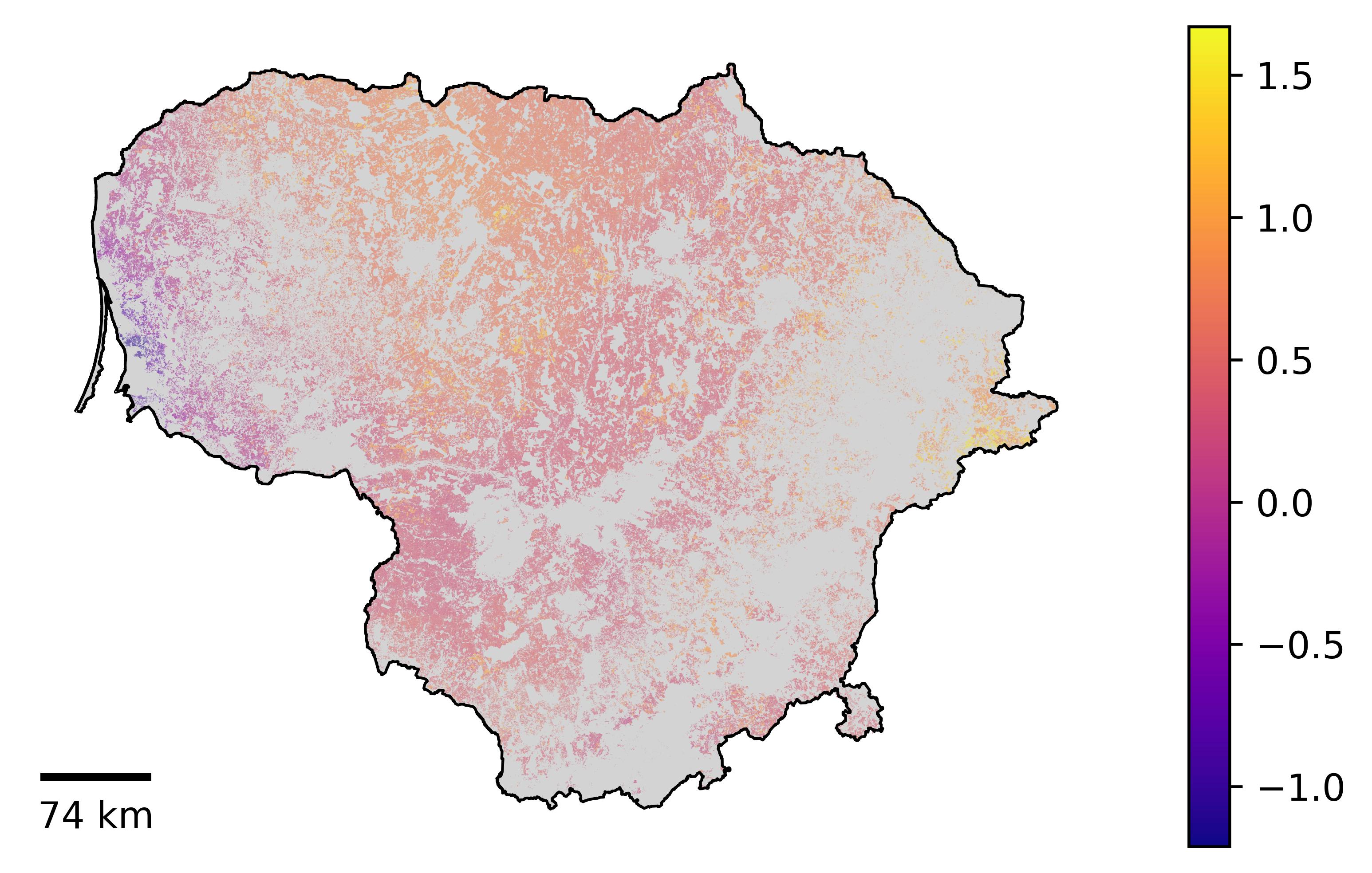}
\caption{Map of the effects (in g/kg) of winter wheat-based rotations on SOC content at the field-level in Lithuania. Maps showing the spatial distribution of key environmental variables and the projected effects for future climate scenarios are provided in the Supplementary Material.}
\label{fig: effect wheat}
\end{figure}

The above findings are further illustrated in the map of Fig. \ref{fig: effect wheat}. In the warmer and wetter areas of western Lithuania, we observe negative or near-zero gains from winter wheat-based rotations. This may be attributed to the high microbial activity in warmer, moister conditions, which accelerates the breakdown of residues or possibly from water logging and/or erosion resulting from frequent or heavy rainfall. In contrast, the Eastern region, where precipitation is relatively low and temperatures are cooler, shows the largest average treatment effects from winter wheat-based rotations. In these zones, residue decomposition proceeds more slowly, leading to more stable but gradual SOC accumulation over time. Overall, temperature and precipitation exert a strong control on SOC because they govern the balance between residue inputs and carbon losses via microbial mineralization and lateral export (e.g., runoff). Global syntheses have shown that higher temperatures accelerate heterotrophic respiration, increasing SOC turnover even when plant inputs rise, indicating sizable soil-C losses under warming \cite{davidson2006temperature}. Precipitation amplifies these thermal effects by maintaining soil water near microbial optima; in warm, wet conditions, decomposition rates peak, whereas drier or cooler conditions suppress respiration. In addition, frequent or intense rainfall increases SOC losses through erosion and hydrologic export of dissolved carbon from soil surface horizons. Thus, in winter wheat–based rotations, cooler/drier areas allow residue-driven gains to accumulate, while warmer/wetter areas push systems toward carbon neutrality (or loss) as faster mineralization and hydrologic and/or erosional pathways offset inputs. Framed within broader soil-carbon management, these patterns echo global guidance that climate constraints can dominate management signals and must be accounted for in site-specific sustainable farming strategies.

Fig. 2 in Supplementary Material shows that the estimated SOC increases due to winter wheat-based rotation are statistically significant primarily in the central and eastern parts of Lithuania, whereas the weaker or negative effects observed in the western regions are not statistically significant. This spatial pattern suggests that while positive SOC responses are robust in certain areas, apparent negative effects may reflect local variability or uncertainty rather than a consistent underlying process.

SOC changes in nearby locations may be more similar than expected by chance due to spatial autocorrelation, meaning that areas close to each other tend to have related values. This spatial dependence can influence the interpretation of localized effects, as clusters of apparently significant SOC changes may partly reflect spatial structure rather than independent effects of winter wheat-based rotations. However, our model includes the latitude and longitude of each field centroid among the control variables ($W$), thereby accounting for spatial variation in environmental and management conditions and reducing potential bias due to spatial location. While including coordinates does not fully eliminate spatial autocorrelation, it helps mitigate location-driven confounding. Consequently, the spatial patterns in Fig. \ref{fig: effect wheat} should be interpreted with this in mind. Future work could incorporate explicit spatial causal inference approaches (e.g., \cite{reich_review_nodate, akbari2023spatial}) to directly account for spatial dependence in treatment effects and uncertainty estimates.

\subsection{Climatic scenarios}

To explore how climate change might influence the effect of winter wheat-based rotations on SOC, we incorporated future temperature and precipitation projections from the CMIP6 model, corresponding to the SSP1-2.6, SSP2-4.5, and SSP5-8.5 scenarios. Since our model was trained on standardized climate features (\(X + W\)), we rescaled the SSP-based projections by subtracting the mean and dividing by the standard deviation of ERA5 temperature and precipitation records from 2018–2022. Rescaling ensured consistency between measurement units of historical observations and future inputs.

\begin{figure}[!ht]
\centering
\includegraphics[width=0.75\columnwidth]{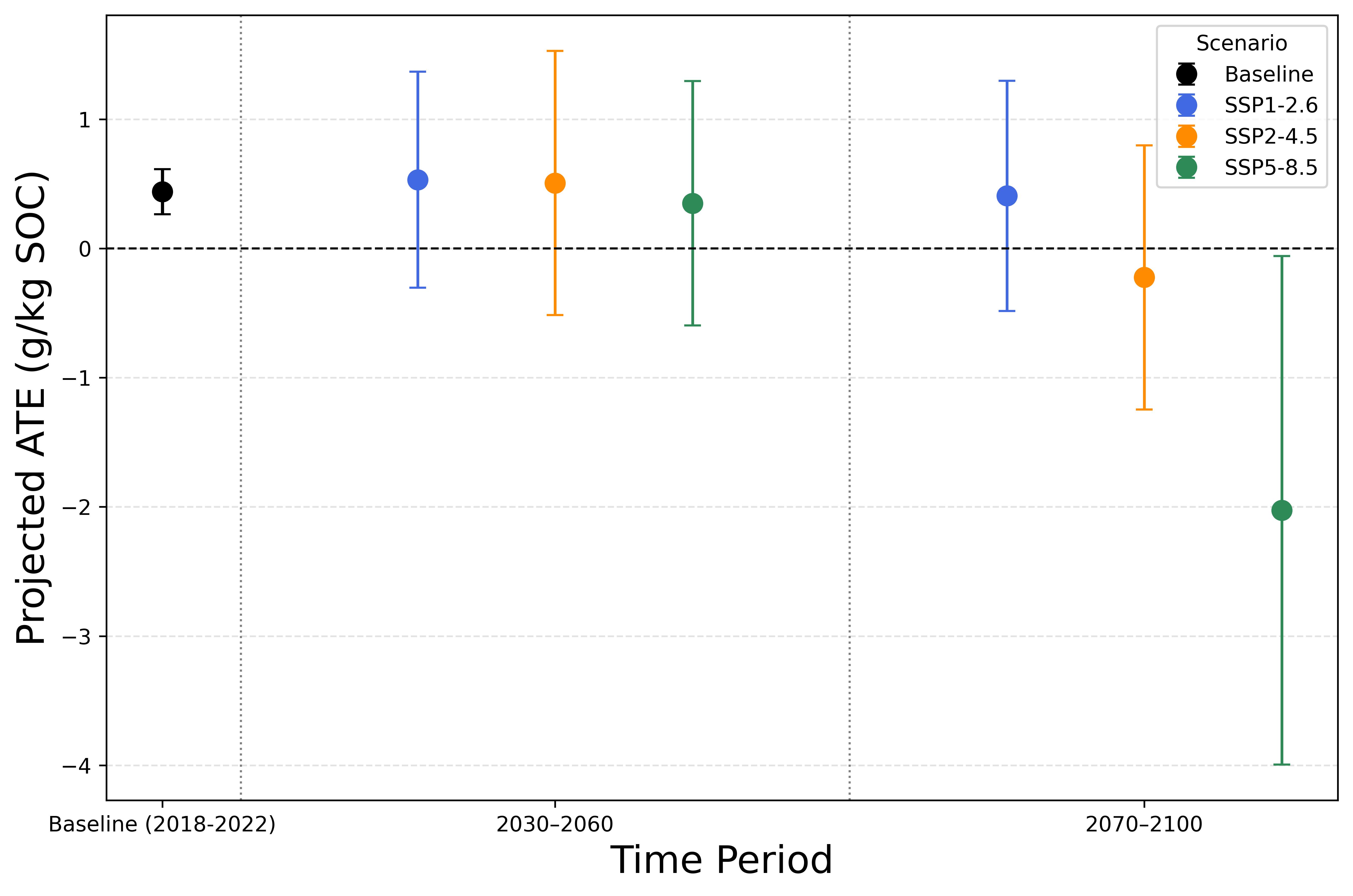}
\caption{Projected ATEs and 95\% confidence intervals of winter wheat-based rotation on SOC under different climate scenarios. The baseline ATE for 2018–2022 is $+0.44$ g/kg SOC. Future projections incorporate SSP-based temperature and precipitation shifts while keeping all other soil and management features constant.}
\label{fig:project}
\end{figure}

Using this adjusted dataset, we estimated the ATE of winter wheat-based rotation on SOC under different SSP scenarios. While all other environmental and soil properties remained constant, temperature and precipitation were updated according to the SSP projections. The results, summarized in Fig. \ref{fig:project}, show how climate-driven changes may affect SOC sequestration potential under various global development pathways. Under the low-emission SSP2.6 scenario, winter wheat maintains a positive SOC effect throughout the century, though with a slight decline from $+0.53$ g/kg SOC in 2030–2060 to $+0.41$ g/kg SOC in 2070–2100. However, under intermediate- and high-emission scenarios, the SOC benefits of wheat rotations lessen over time.  

In SSP4.5, the ATE remains positive in the mid-century ($+0.51$ g/kg SOC) but turns negative ($-0.22$ g/kg SOC) by the late century. This shift is even more pronounced in SSP8.5, where the initial benefit is smaller ($+0.35$ g/kg SOC in 2030–2060) and transitions to a sharp loss ($-2.03$ g/kg SOC) by 2070–2100. These findings indicate that while wheat-based rotations contribute to SOC sequestration, their effectiveness will likely diminish under warming scenarios. The observed trend aligns with expectations that rising temperatures and altered precipitation patterns accelerate SOC decomposition rates, outpacing potential sequestration gains. 

It is important to interpret these projections as indicative of the potential climate-driven effects on SOC under winter wheat-based rotations, rather than precise forecasts of future SOC content. Our estimates assume that all soil properties and management practices remain constant over time, while only temperature and precipitation are adjusted according to SSP scenarios. Consequently, observed trends, such as the decline of SOC benefits under high-emission scenarios, reflect the sensitivity of SOC to climate alone. In reality, soil characteristics, crop management, and other environmental factors may change due to human intervention, technological improvements or feedbacks from climate change, which could mitigate or amplify these effects. This assumption introduces uncertainty into the projections and highlights the need for future studies to incorporate dynamic soil and management processes for more comprehensive assessments. Overall, these findings underscore the potential vulnerability of SOC sequestration to climate change and provide insights into where adaptive management may be necessary to maintain soil carbon stocks.

\subsection{Using CATEs for decision-making}

The impact of winter wheat-based rotations on SOC varies spatially across Lithuania, primarily due to differences in temperature and precipitation. Our CATE analysis reveals that climate moderates treatment effects: in cooler, drier regions, winter wheat-based rotation tend to enhance SOC, whereas in warmer, wetter areas, benefits are diminished by faster residue decomposition and erosion risk.

These findings support region-specific recommendations. For example, winter wheat adoption could be promoted in the cooler eastern regions, while in the warmer west, alternative practices such as cover cropping or erosion control may be more effective. Tailoring management strategies to local climate conditions can improve both environmental outcomes and agronomic performance. CATE estimates can also inform policy design, enabling regionally targeted incentives and conservation programs that reflect where winter wheat-based rotation are most beneficial. This alignment of science and policy supports more efficient and equitable resource use.

Climate change further affects these dynamics. Under a low-emission scenario (SSP2.6), SOC benefits from winter wheat persist, albeit with a modest decline. Under a high-emission scenario (SSP8.5), those benefits erode significantly, with SOC loss becoming a concern later in the century. This highlights the urgency of adaptive management strategies that are both region-specific and climate-resilient.

By leveraging CATE-informed insights, farmers and policymakers can adopt precision interventions that maximize SOC gains and maintain long-term sustainability under shifting environmental conditions.

\section{Conclusions}

Our findings demonstrate that winter wheat-based rotations can enhance SOC, with an average effect of +0.44 g/kg SOC across Lithuania. Importantly, increased SOC is not inherently incompatible with intensive agriculture: higher winter wheat use, often considered an indicator of agricultural intensity, is associated with SOC gains, suggesting that carefully designed intensive rotation regimes can contribute to climate change mitigation. However, these benefits are not uniform across space. Cooler regions show stronger SOC gains, whereas warmer areas exhibit diminished or even negative effects, indicating SOC outcomes can be optimized by favoring wheat-based rotations in cooler regions and adopting alternative SOC-friendly practices in warmer zones.

When looking ahead, climate projections reveal that these spatially heterogeneous benefits may also vary over time. Under high-emission scenarios, SOC gains are expected to weaken and may even reverse by the end of the century, with the ATE declining to –2.03 g/kg SOC under SSP5–8.5. This highlights the sensitivity of SOC sequestration to changing temperature and precipitation regimes. For policymakers, these projected declines underscore the need for region-specific adaptation strategies, supporting SOC-enhancing rotations where potential remains high and encouraging alternative practices in areas at risk of future SOC losses.

By applying causal machine learning, we demonstrate how explicitly modeling both spatial and climate-driven heterogeneity can guide targeted, climate-resilient soil carbon management strategies.

\begin{ack}
    This work has been supported by the EIFFEL project (EU Horizon 2020 - GA No. 101003518) and the GVA PROMETEO AI4CS Project (CIPROM/2021/056).
\end{ack}

\section*{Data availability statement}
The high resolution soil layers at national scale or Lithuania (2020-2022) are available at https://gkhub.earthobservations.org/records/11d3n-nv143. The LPIS (parcel-level) data can be found at https://www.geoportal.lt/geoportal/en/web/en/. CATE model results are provided in the following interactive visualization https://portfolio.i-bec.org/eiffel/ which includes a download functionality.

\section*{Conflicts of Interest}
The authors declare no conflicts of interest.

\end{document}